\documentclass[journal]{IEEEtran}
\usepackage{float}

\usepackage[utf8]{inputenc} % 编码
\usepackage{amsmath,amsfonts,amssymb} % 数学公式支持
\usepackage{tabularx} % 导入宏包
\usepackage{graphicx} % 图片
\usepackage{cite} % 引用格式
\usepackage[linesnumbered,ruled,vlined]{algorithm2e}
\usepackage{graphicx} % 确保导入宏包
 
\usepackage{graphicx} % Required for inserting images
\usepackage{stfloats}
\usepackage{graphicx}
\usepackage{cite}
\usepackage[caption=false,font=footnotesize]{subfig}
\usepackage{bm}   % 用于加粗数学符号
\title{Recursive Vision Transformer with Dynamic Depth and Width Adjustment for Resource-Efficient Image Semantic Communication}
\author{Zhilong Zhang~\IEEEmembership{Member,~IEEE}, 
Xinhui Zhang, 
Gongyu Jin, 
Sihua Wang~\IEEEmembership{Student Member,~IEEE},\\ 
Danpu Liu~\IEEEmembership{Senior Member,~IEEE}, 
and Changchuan Yin~\IEEEmembership{Senior Member,~IEEE}%
\thanks{This work is supported by National Natural Science Foundation of China with grant 62271065 and U22B2001.}%
\thanks{Zhilong Zhang, Xinhui Zhang, Gongyu Jin, Sihua Wang, Danpu Liu, and Changchuan Yin are
with Beijing Laboratory of Advanced Information Network, and the Beijing
Key Laboratory of Network System Architecture and Convergence, Beijing
University of Posts and Telecommunications, Beijing 100876, China
(e-mail: zhangzhilong@bupt.edu.cn, zhangxinhui@bupt.edu.cn, jingongyu@gz.chinamobile.com,
sihuawang@bupt.edu.cn, dpliu@bupt.edu.cn, ccyin@bupt.edu.cn).}%
\thanks{Corresponding Author: Sihua Wang.}%
}

\begin{document}

\maketitle

\begin{abstract}
Image semantic communication is a critical component in next-generation wireless communication systems. However, such systems typically suffer from large memory footprints and high computational complexity, making them difficult to deploy on resource-constrained devices. To address these challenges, we propose a vision transformer (ViT)–enabled image semantic communication system. In this system, a recursive structure is introduced to iteratively refine semantic features and reduce the parameter count. In addition, three dynamic adjustment strategies are designed to adaptively reduce computational complexity: dynamic depth adjustment, dynamic width adjustment, and joint width–depth optimization. Dynamic depth adjustment adaptively determines the number of recursive modules according to image content and channel conditions, while dynamic width adjustment selectively preserves important neurons and attention heads. The joint width–depth optimization further enables flexible computation configurations. Simulation results verify that the proposed recursive ViT–based system, combined with the three dynamic adjustment strategies, reduces the parameter count by 48.7\% and achieves higher reconstruction quality than existing baselines under comparable computational complexity.
\end{abstract}

\begin{IEEEkeywords}
Image semantic communication, recursive ViT, dynamic depth adjustment, dynamic width adjustment, joint width–depth optimization.
\end{IEEEkeywords}

\section{Introduction} 
Images play an important role in future communication as powerful carriers of rich information, supporting a wide range of critical applications such as healthcare, entertainment, and intelligent transportation~\cite{ref32,ref33,ref34}. Nevertheless, the massive volume and high resolution of image data impose substantial burdens on conventional communication systems, resulting in excessive bandwidth usage and limited transmission efficiency. To address these limitations, researchers have turned their attention to semantic communication, a key enabling technology for sixth-generation (6G) communications that shifts the emphasis from traditional bit-level transmission to the conveyance of the underlying semantic information of the data~\cite{ref1,ref2,ref3,ref4,ref15}. Existing studies commonly employ deep learning models to extract semantic features of images in an end-to-end manner. However, such models typically exhibit large memory footprints and high computational complexity, which restrict their deployment on resource-constrained platforms such as Internet of Things (IoT) devices.

In response, recent studies such as~\cite{ref6,ref7,ref8,ref9} focused on developing lightweight image semantic communication solutions that achieve strong performance under strict resource constraints. In \cite{ref6}, the authors added a lightweight attention module to each downsampling layer so that fewer layers are required while still ensuring effective semantic extraction. In~\cite{ref7}, depthwise separable convolutions are employed in the semantic encoder and decoder, where standard convolutions are decomposed into depthwise and pointwise operations, reducing the number of convolutional parameters and multiply-accumulate operations. The authors in~\cite{ref8} adopted a pruning strategy based on the $\ell_{1}$-norm regularization, which removes less important parameters. In~\cite{ref9}, a fast distillation method with a pre-stored compression mechanism was proposed to reduce computational complexity. However, the parameter-reduction ratios on learning models of these lightweight approaches are fixed once training is completed, leaving no flexibility to adapt during inference. As a result, they struggle to respond to input samples and wireless channel conditions, often leading to inefficient computation and unstable reconstruction quality. These limitations highlight the need for more flexible approaches capable of adaptive computation.

To address this issue, one feasible approach is dynamic neural networks (DNNs)~\cite{ref10}, which can adjust the structure of the deep learning model in the semantic communication framework during inference according to input samples and wireless environments. In~\cite{ref30}, the authors employed a dynamic strategy to adjust active feature dimensions according to channel conditions. Similarly, the authors in~\cite{ref31} adopted an adaptive preprocessing method that selectively masks less important image regions based on their content and channel conditions, prioritizing critical parts during transmission. However, these works mainly focus on transmission reliability rather than reducing the internal computational complexity. The potential of DNNs to improve computational efficiency in semantic communication remains to be further explored.

While DNNs can alleviate computational redundancy during inference, they still require storing all model parameters. To address this challenge, recursive architectures offer a complementary solution~\cite{ref20}. Instead of stacking numerous distinct layers, a recursive architecture repeatedly applies shared modules across multiple stages, effectively reducing the number of unique parameters while maintaining the expressive capacity of the semantic encoder and decoder. Motivated by these challenges, this study investigates image semantic communication systems by combining recursive architectures with DNNs, aiming to reduce parameter redundancy and enhance computational efficiency. Our key contributions are summarized below: 
\begin{itemize} 
\item \textbf{Parameter-Efficient Recursive Transformer Design for Memory Saving.} ViT~\cite{ref12} can effectively extract essential semantic features, which is crucial for image semantic communication. However, stacking multiple Transformer layers greatly increases parameter count. Although recursion offers an effective strategy to alleviate this issue, its efficiency often comes at the cost of parameter sharing which may limit representational diversity. To overcome this limitation, we propose a recursive Transformer design that employs multiple Transformer encoder layers as independent recursive modules. Within each module, the same Transformer encoder layer is recursively applied twice. This design leverages recursion for parameter efficiency while maintaining representational diversity.

\item \textbf{Content-Aware Feature Extraction for Low Computational Complexity.} 
The computational complexity of semantic feature extraction might be different across images. 
To obtain satisfactory semantic information with minimal complexity, we propose a stopping-score-based dynamic depth adjustment strategy, allowing the inference process to terminate early once sufficient semantic features have been extracted. 
Specifically, we implement a cumulative stopping score mechanism, which maps the feature changes at each layer into a comparable scalar to decide whether to continue computation. 
The mechanism incorporates a loss function that jointly accounts for reconstruction quality and computational complexity, enabling reduced computation while preserving high reconstruction quality.

\item \textbf{Channel-Aware Adaptive Computation for Robust Transmission.} In image semantic communication scenarios, different channel conditions can affect reconstructed image quality. This motivates us to extend the cumulative stopping score mechanism by incorporating channel conditions into the depth decision process. In addition, we propose a dynamic width adjustment strategy via pruning, which adaptively modulates the width according to channel conditions. This strategy consists of two main components. First, we embed the channel conditions into a differentiable threshold, allowing continuous channel variations to smoothly guide discrete pruning decisions during training, thereby achieving adaptive width allocation. Second, we introduce module-specific adaptive thresholds, enabling each module to fine-tune its pruning sensitivity based on feature importance, thereby preserving critical features while minimizing redundant computation.
\end{itemize} 

The remainder of this paper is organized as follows. Section II details the design of the recursive ViT–based semantic communication system. Section III introduces the proposed dynamic optimization strategies. Section IV presents the simulation results and evaluation. Finally, Section V concludes this paper.

\section{System Model}

\begin{figure*}[t]
    \centering
    \includegraphics[clip, trim=0.1cm 0.1cm 0.1cm 0.1cm, width=\textwidth]{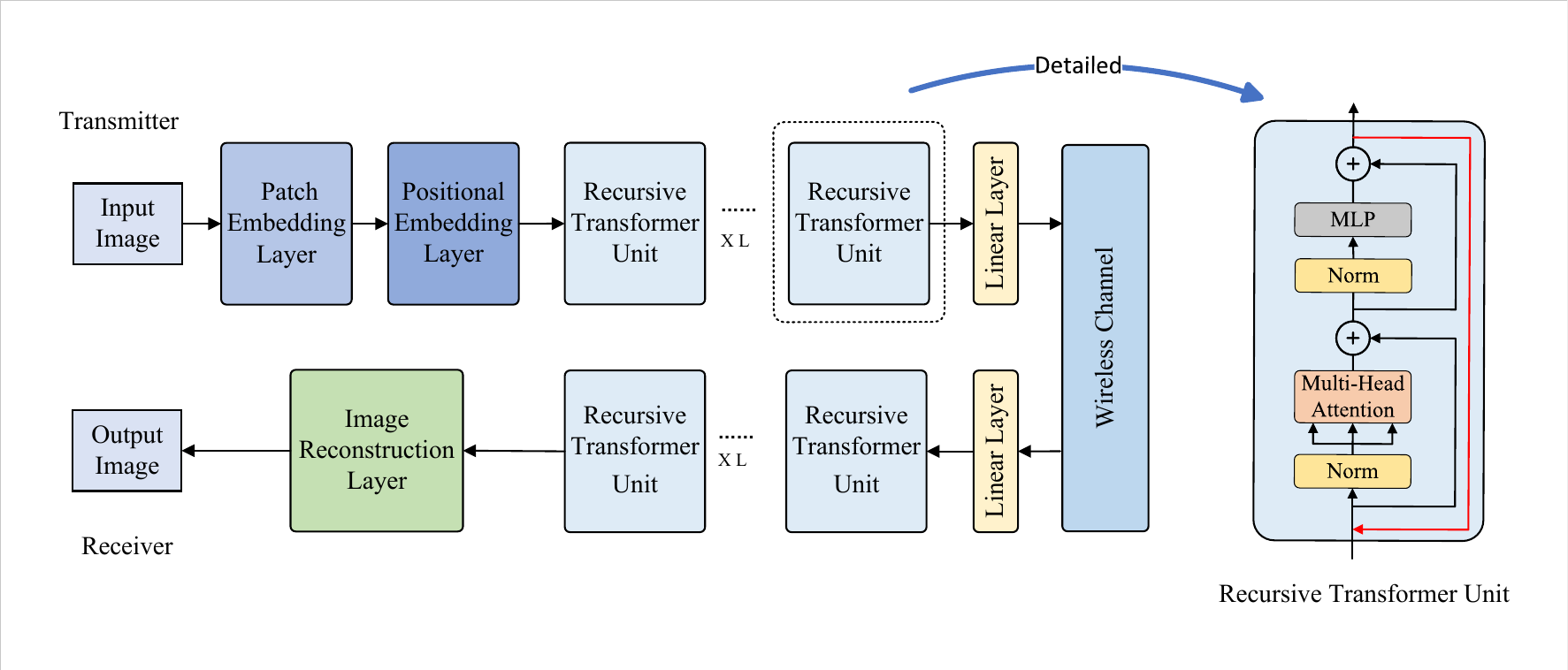}
    \caption{Illustration of the proposed recursive ViT–based image semantic communication system.}
    \label{fig:enc_dec}
\end{figure*}

\subsection{ViT–Based Image Semantic Communication Framework}

Image semantic communication aims to achieve efficient and robust image transmission by encoding semantic-level features instead of raw pixel data.
In ViT-based image semantic communication systems~\cite{ref18,ref19}, the processing pipeline generally consists of three main stages: image feature extraction, image feature transmission, and image reconstruction.

\subsubsection{Image Feature Extraction}

This stage converts the input image into compact and meaningful semantic representations through three steps: patch embedding, positional encoding, and Transformer encoding.   

\textbf{Patch Embedding:} 
Patch embedding converts the input image into a sequence of vectorized patches. 
Let $\bm{s} \in \mathbb{R}^{H_s \times W_s \times C}$ denote the input image, 
where $H_s$, $W_s$, and $C$ represent the spatial height, spatial width, and number of channels, respectively.
This step can be written as
\begin{equation}
\mathbf{Z}_{\mathrm{patch}}= f_p(\bm{s}) \in \mathbb{R}^{M \times E},
\end{equation}
where $\bm{s}$ is divided into $M$ non-overlapping $P\times P$ patches and each patch is flattened into a vector of dimension $E=P\times P\times C$.

\textbf{Positional Encoding:} 
Positional encoding injects spatial information to preserve patch locations. A positional embedding function $f_{pos}(\cdot)$ is applied to encode spatial information, yielding the initial patch embeddings:
\begin{equation}
\mathbf{Z}^{(0)} = \mathbf{Z}_{\text{patch}} + f_{pos}(\mathbf{Z}_{\text{patch}}).
\end{equation}

\textbf{Transformer Encoding:} 
Transformer encoding is performed through $L$ Transformer encoder layers, each consisting of a multi-head self-attention (MHSA) module and a feed-forward multi-layer perceptron (MLP) module~\cite{ref14}. This step produces the final semantic representation across patches.
Let $\mathbf{Z}^{(l-1)} \in \mathbb{R}^{M \times E}$ denote the input to the $l$-th Transformer encoder layer. 
For each attention head $i = 1, \dots, H$, the queries, keys, and values are obtained by linear projections:
\begin{align} \mathbf{Q}_i^{(l)} &= \mathbf{Z}^{(l-1)} \mathbf{W}_i^{Q,(l)}, \\ \mathbf{K}_i^{(l)} &= \mathbf{Z}^{(l-1)} \mathbf{W}_i^{K,(l)}, \\ \mathbf{V}_i^{(l)} &= \mathbf{Z}^{(l-1)} \mathbf{W}_i^{V,(l)}, \end{align}
where $\mathbf{W}_i^{Q,(l)}, \mathbf{W}_i^{K,(l)}, \mathbf{W}_i^{V,(l)} \in \mathbb{R}^{E \times d_k}$ are layer-specific learnable projection matrices.

The query, key, and value matrices are then used to compute the attention output for each head, capturing contextual dependencies among patches via scaled dot-product attention:
\begin{equation}
\mathbf{head}_i^{(l)} = \text{Softmax}\left( \frac{\mathbf{Q}_i^{(l)} (\mathbf{K}_i^{(l)})^\top}{\sqrt{d_k}} \right) \mathbf{V}_i^{(l)},
\end{equation}
where $d_k = E / H$ denotes the dimensionality of each attention head.
 
The outputs from all attention heads are concatenated and projected:
\begin{equation}
\text{MHSA}^{(l)}(\mathbf{Z}^{(l-1)}) = \text{Concat}(\mathbf{head}_1^{(l)}, \dots, \mathbf{head}_H^{(l)}) \mathbf{W}^{O,(l)},
\end{equation}
where $\mathbf{W}^{O,(l)} \in \mathbb{R}^{H d_k \times E}$ is the output projection matrix that linearly maps the concatenated multi-head features into the embedding space of dimension $E$.
Subsequently, a residual connection and layer normalization are applied:
\begin{equation}
\mathbf{Z}'^{(l)} = \mathbf{Z}^{(l-1)} + \mathrm{MHSA}^{(l)}\!\left(
\mathrm{LN}_{1}^{(l)}\!\left(\mathbf{Z}^{(l-1)}\right)\right).
\end{equation}

Next, an MLP module is applied, followed by another residual connection and normalization:
\begin{equation}
\mathbf{Z}^{(l)} = \mathbf{Z}'^{(l)} + \mathrm{MLP}^{(l)}\!\left(
\mathrm{LN}_{2}^{(l)}\!\left(\mathbf{Z}'^{(l)}\right)\right),
\end{equation}
where the MLP is defined as
\begin{equation}
\mathrm{MLP}^{(l)}(\mathbf{X}) = \mathbf{W}_2^{(l)} \, \mathrm{sigmoid}(\mathbf{W}_1^{(l)} \mathbf{X} + \bm{b}_1^{(l)}) + \bm{b}_2^{(l)},
\end{equation}
with $\mathbf{W}_1^{(l)} \in \mathbb{R}^{E \times D}$, $\mathbf{W}_2^{(l)} \in \mathbb{R}^{D \times E}$, and $\bm{b}_1^{(l)}$ and $\bm{b}_2^{(l)}$ serving as layer-specific learnable parameters, where $D$ denotes the MLP hidden dimension and $\mathrm{sigmoid}(\cdot)$ denotes the activation function.

After processing the input patch embeddings through $L$ Transformer encoder layers, 
the resulting global semantic representation is denoted as $\mathbf{Z}^{(L)}$:
\begin{equation}
\mathbf{Z}^{(L)} = f_s(\mathbf{Z}^{(0)}),
\end{equation}
where $f_s(\cdot)$ denotes the complete semantic encoding function that transforms the initial patch embeddings $\mathbf{Z}^{(0)}$ into the final semantic representation.

\subsubsection{Image Feature Transmission}

The encoded semantic features $\mathbf{Z}^{(L)}$ are mapped to channel symbols through the joint source–channel encoder $f_c(\cdot)$:
\begin{equation}
\bm{x} = f_c(\mathbf{Z}^{(L)}).
\end{equation}

Then, the feature vector $\bm{x}$ is transmitted through a wireless channel,
which is
\begin{equation}
\bm{y} = g \bm{x} + \bm{n},
\end{equation}
where $g \in \mathbb{C}$ denotes the complex fading coefficient, and 
$\bm{n} \sim \mathcal{CN}(\mathbf{0}, \sigma^2 \mathbf{I})$ represents the additive white Gaussian noise (AWGN) vector.

\subsubsection{Image Reconstruction}

At the receiver side, the channel decoder $f_c^{-1}(\cdot)$ and semantic decoder $f_s^{-1}(\cdot)$ are sequentially applied to recover the semantic features. 
The reconstructed features are then processed by the image reconstruction module $f_r(\cdot)$ to generate the final output:
\begin{equation}
\hat{\bm{s}} = f_r\Big(f_s^{-1}\big(f_c^{-1}(\mathbf{y})\big)\Big).
\end{equation}

The entire framework is optimized end-to-end using the mean squared error (MSE) loss function:
\begin{equation}
\mathcal{L}_{\mathrm{MSE}} = \frac{1}{N} \sum_{i=1}^{N} \| \bm{s}_i - \hat{\bm{s}}_i \|_2^2,
\end{equation}
where $N$ is the total number of training samples.

\subsection{Recursive ViT–Based Image Semantic Communication System}

The ViT provides powerful global modeling and feature representation capabilities by leveraging self-attention mechanisms. However, these advantages come at the cost of substantial parameter overhead and high memory consumption. A standard Transformer encoder layer consists of an MHSA module and an MLP module.
In the MHSA module, four projection matrices with size $E \times E$ result in roughly $4E^{2}$ learnable parameters, whereas the MLP module includes two weight matrices with size $E \times D$ and $D \times E$, yielding about $2ED$ parameters in total.
For $L$ stacked layers, the total parameter count can be approximated as
$L(4E^2 + 2ED)$.
As the parameters grow with $L$, $E$, and $D$, the overall memory requirement becomes prohibitive for deployment on resource-constrained semantic communication devices.

To reduce parameter count while maintaining semantic expressiveness, we propose a Recursive Transformer Unit (RTU), as illustrated in Fig.~\ref{fig:enc_dec}. 
Each RTU consists of a single Transformer encoder layer whose output is recursively fed back as input. 
Recursion enables the RTU to achieve deeper semantic representations with fewer parameters. However, repeatedly using the same parameters may limit the refinement of features.
To balance parameter efficiency and representational richness, we apply the encoder twice within each RTU. 
Specifically, each RTU recursively applies its Transformer encoder layer in two consecutive steps, which are described as follows:

\textbf{Step 1 — First recursion of the $k$-th RTU:} 
The first recursion aims to extract high-level semantic features from the input while preserving the original information.
The input $\mathbf{Z}^{(2k-2)}$ is fed into an MHSA module to capture global dependencies among patches, followed by a residual connection and layer normalization:
\begin{equation}
\mathbf{Z}'^{(2k-1)} = \mathbf{Z}^{(2k-2)} + \mathrm{MHSA}^{(k)}\!\left(
\mathrm{LN}_1^{(k)}\!\left(\mathbf{Z}^{(2k-2)}\right)\right).
\end{equation}

The intermediate output is then refined via an MLP module with residual connection and layer normalization:
\begin{equation}
\mathbf{Z}^{(2k-1)} = \mathbf{Z}'^{(2k-1)} + \mathrm{MLP}^{(k)}\!\left(
\mathrm{LN}_2^{(k)}\!\left(\mathbf{Z}'^{(2k-1)}\right)\right).
\end{equation}

\textbf{Step 2 — Second recursion of the $k$-th RTU:}  
The second recursion aims to reinforce feature consistency and consolidate the semantic representation obtained from the first recursion.
The output from Step 1 is fed into the MHSA module with residual connection and normalization, using the same set of parameters:
\begin{equation}
\mathbf{Z}'^{(2k)} = \mathbf{Z}^{(2k-1)} + \mathrm{MHSA}^{(k)}\!\left(
\mathrm{LN}_1^{(k)}\!\left(\mathbf{Z}^{(2k-1)}\right)\right),
\end{equation}
followed by the MLP module with residual connection and normalization:
\begin{equation}
\mathbf{Z}^{(2k)} = \mathbf{Z}'^{(2k)} + \mathrm{MLP}^{(k)}\!\left(
\mathrm{LN}_2^{(k)}\!\left(\mathbf{Z}'^{(2k)}\right)\right).
\end{equation}

By sequentially applying $L/2$ RTUs, the same effective depth as $L$ Transformer encoder layers is achieved.
This results in a 50\% reduction in parameters while preserving semantic representation. 
The RTU serves as the core module for both the semantic encoder and decoder, reducing memory requirements while enabling deployment on resource-constrained devices.

\section{Dynamic Adjustment Strategies for the Recursive ViT-Based Image Semantic Communication System}

The proposed RTU effectively reduces the parameter count.
However, the recursive ViT–based semantic communication system still suffers from relatively high computational complexity during inference, which is commonly measured in terms of floating-point operations (FLOPs). 
To identify the main sources of computational complexity, we analyze the FLOPs of a Transformer encoder
layer, as detailed in Table~\ref{tab:transformer_complexity}.

From Table~\ref{tab:transformer_complexity}, it is clear that the total FLOPs grow with $L$, $E$, $M$, and $D$. However, the actual computational complexity varies with the difficulty of semantic feature extraction from the input image and the channel conditions. Therefore, to adapt to such variations while maintaining high reconstruction quality, we propose three dynamic adjustment strategies that reduce computational complexity from different aspects.\footnote{The proposed strategies are applied to both the encoder and decoder, each containing $L/2$ RTUs, which correspond to $L$ Transformer encoder layers. Since their structures are similar, the encoder is used as an example in the following description.}

\subsection{Dynamic Depth Adjustment Strategy}

To enable dynamic depth control during inference, we adopt a dynamic depth adjustment strategy using stopping scores, which allows each layer to decide whether a patch requires further processing or not, thereby reducing unnecessary computation. The following description is organized in two steps for clarity.

\begin{table}[t]
\centering
\caption{FLOPs of Each Component in a Transformer Encoder Layer}
\label{tab:transformer_complexity}
\resizebox{0.48\textwidth}{!}{
\begin{tabular}{ll}
\hline
\textbf{Component} & \textbf{FLOPs} \\
\hline
Linear projections for $Q,K,V$ & $3M E^2$ \\
Attention score computation & $M^2 E/h$ \\
Weighted value aggregation & $M^2 E$ \\
Output projection & $M E^2$ \\
\textbf{Total MHSA} & $4M E^2 + 2M^2 E$ \\
\hline
Two linear layers (MLP) & $2MED$ \\
\textbf{Total MLP} & $2MED$ \\
\hline
\textbf{Total per layer} & $4M E^2 + 2M^2 E + 2MED$ \\
\hline
\end{tabular}}
\end{table}

\textbf{Step 1 — Patch-wise Feature Update and Stopping-State Computation:}  
At layer \(l\), let \(\bm{p}^{(l-1)} \in \{0,1\}^M\) denote the binary indicator vector from the previous layer. 
For each patch \(m = 1, \dots, M\), 
\(p_m^{(l-1)} = 1\) indicates that the patch is active and will be processed at layer \(l\), 
and \(p_m^{(l-1)} = 0\) otherwise.
According to the stop decisions \({p_m^{(l-1)}}\), each patch is updated as

\begin{equation}
\bm{z}_m^{(l)} =
\begin{cases}
f_s^{(l)}(\bm{z}_m^{(l-1)}), & p_m^{(l-1)} = 1,\\[2mm]
\bm{z}_m^{(l-1)}, & p_m^{(l-1)} = 0,
\end{cases}
\end{equation}
where $f_s^{(l)}(\cdot)$ denotes the function of the $l$-th Transformer encoder layer.

For the active patches, we define a stop-score vector \(\bm{h}^{(l)} \in \mathbb{R}^M\), where each element quantifies the contribution of the corresponding patch at the current layer:

\begin{equation}\label{stopsores}
h_m^{(l)} =
\begin{cases}
\mathrm{sigmoid}\!\Big(\alpha (\bm{w}^\top \bm{z}_m^{(l)} + \bm{b}) - \beta \sigma^2 \Big), & p_m^{(l-1)} = 1,\\[1mm]
0, & p_m^{(l-1)} = 0,

\end{cases}
\end{equation}
where \(\bm{w} \in \mathbb{R}^{E}\) and \(\bm{b}\) are learnable parameters, 
$\alpha$ and $\beta$ are hyperparameters, and \(\sigma^2\) represents the channel condition factor.

To accumulate the historical contribution of each patch across layers, we maintain a cumulative score vector 
\(\tilde{\bm{h}}^{(l)} \in \mathbb{R}^M\), where each element is updated as

\begin{equation}
\tilde{h}_m^{(l)} = \tilde{h}_m^{(l-1)} + h_m^{(l)}.
\end{equation}

Based on the cumulative score, the binary indicator vector for the current layer is updated:

\begin{equation}
p_m^{(l)} =
\begin{cases}
0, & \tilde{h}_m^{(l)} \ge T, \\
1, & \text{otherwise},
\end{cases}
\end{equation}
where \(T\) is the stopping threshold.

\textbf{Step 2 — Layer-wise Feature Weighting and Aggregation:}  
Given the patch features $\{\bm{z}_m^{(l)}\}_{m=1}^M$ computed by Eq.~(20), we collect them into the feature matrix:

\[
\mathbf{Z}^{(l)} = 
\begin{bmatrix}
\bm{z}_1^{(l)}, \dots, \bm{z}_M^{(l)}
\end{bmatrix}^\top
\in \mathbb{R}^{M \times E}.
\]

Different patches contribute unequally to the layer output. To account for the residual contribution of patches that have stopped, we define a remaining score vector \(\bm{r}^{(l)} \in \mathbb{R}^M\), with each element computed as
\begin{equation}
r_m^{(l)} = 1 - \tilde{h}_m^{(l-1)}.
\end{equation}

The output weight for each patch is computed by combining the stop score and the remaining score:

\begin{equation}
\bm{w}^{(l)} = \bm{h}^{(l)} \odot \bm{p}^{(l-1)} + \bm{r}^{(l)} \odot (\mathbf{1} - \bm{p}^{(l-1)}),
\end{equation}
where \(\odot\) denotes element-wise multiplication.

The final output is obtained by summing the weighted features across all layers:
\begin{equation}
\mathbf{Z}_\text{out} = \sum_{l=1}^{L^*} \mathrm{diag}(\bm{w}^{(l)}) \, \mathbf{Z}^{(l)},
\end{equation}
where $L^*$ denotes the effective maximum depth across all patches.

Based on the above mechanism, each patch adaptively determines its computation depth, as summarized in Algorithm~1.
To guide this process, the loss function jointly considers reconstruction quality and computational efficiency.

The computation of all patches is tracked across layers using a counter vector $\bm{q}^{(l)} \in \mathbb{Z}_{\ge 0}^M$, where each element keeps a running total for one patch:
\begin{equation}
q_m^{(l)} = q_m^{(l-1)} + p_m^{(l-1)}.
\end{equation}

Since $q_m^{(l)}$ is discrete and non-differentiable, we introduce the remaining score $r_m^{(l)}$ as its continuous proxy. 
The overall ponder cost is then defined as
\begin{equation}
\mathcal{L}_{\text{ponder}} = \frac{1}{M} \sum_{m=1}^{M} \big( q_m^{L^*} + r_m^{L^*} \big),
\end{equation}
which measures the expected computation per patch while remaining differentiable for gradient-based optimization.

Finally, the overall objective combines reconstruction quality and computation efficiency as
\begin{equation}
    \mathcal{L} = \mathcal{L}_{\text{MSE}} + \tau \, \mathcal{L}_{\text{ponder}},
\end{equation}
where $\tau$ is a trade-off factor.

\begin{algorithm}[htbp]
\caption{Dynamic Depth Algorithm Based on Stopping Scores}
\KwIn{Patch features $\mathbf{Z}$, stop threshold $T \in (0,1)$}
\KwOut{Final output $\mathbf{Z}_\text{out}^*$, cumulative counts $\bm{q}^*$, remaining scores $\bm{r}^*$, stopping layer $L^*$}

Initialize $\bm{p} \gets \bm{1}$, $\bm{q} \gets \bm{0}$, $\bm{r} \gets \bm{1}$, $\tilde{\bm{h}} \gets \bm{0}$, $\mathbf{Z}_\text{out} \gets \bm{0}$, $l \gets 0$\;

\While{any($\tilde{\bm{h}} < T$)}{
    $l \gets l+1$\;

    Update computation count: 
    $\bm{q} \gets \bm{q} + \bm{p}$\;

    Compute patch update: 
    $\mathbf{Z} \gets \bm{p} \odot f_s^{(l)}(\mathbf{Z}) + (\mathbf{1}-\bm{p}) \odot \mathbf{Z}$ 

    \eIf{$l \neq L$}{
        Update stop scores $\bm{h}$ according to Eq.~\eqref{stopsores}
        
    }{
        $\bm{h} \gets \bm{1}$\;
    }

    Update remaining score: 
    $\bm{r} \gets \bm{1} - \tilde{\bm{h}}$\;

    Update cumulative score: 
    $\tilde{\bm{h}} \gets \tilde{\bm{h}} + \bm{h}$\;

    Compute output weight: 
    $\bm{w} \gets \bm{h} \odot \bm{p} + \bm{r} \odot (\bm{1}-\bm{p})$\;

    Update patch indicator $\bm{p}$\;
    
    Update overall output: 
    $\mathbf{Z}_\text{out} \gets \mathbf{Z}_\text{out} + \mathrm{diag}(\bm{w}^{(l)}) \, \mathbf{Z}^{(l)}$
}

\Return $\mathbf{Z}_\text{out}^* \gets \mathbf{Z}_\text{out},\; \bm{q}^* \gets \bm{q},\; \bm{r}^* \gets \bm{r},\; L^* \gets l$\;
\end{algorithm}

\subsection{Dynamic Width Adjustment Strategy}

As shown in Table~\ref{tab:transformer_complexity}, the majority of FLOPs in each layer come from operations on the weight matrices.
To reduce computational complexity, we propose a dynamic width adjustment strategy that selectively prunes less important units in the weight matrices according to channel conditions.

We denote the weight matrix by 
$\mathbf{W}$
and use a binary mask matrix 
$\mathbf{M}$
to determine the active units during the forward pass.
Specifically, $\mathbf{M}$ is obtained by expanding a binary vector 
$\bm{m} = [m_1, \dots, m_n]^\top \in \{0,1\}^n$, with $n$ denoting the total number of prunable units in $\mathbf{W}$. Each entry $m_i \in \{0,1\}$ corresponds to a neuron or an attention head, with $m_i = 1$ indicating retention and $m_i = 0$ indicating pruning. Then, the pruned weight matrix is defined as
\begin{equation}
\mathbf{W}_{\mathrm{pruned}} = \mathbf{M} \odot \mathbf{W}.
\label{eq:W_pruned}
\end{equation}

The binary vector $\bm{m}$ is determined based on a learnable importance score vector $\bm{a} = [a_1, \dots, a_n]^\top$ and an adaptive threshold $\xi$~\cite{ref21}.  Specifically, each entry $m_i$ is computed as
\begin{equation}
m_i =
\begin{cases}
1, & i \in \mathrm{sort}(\bm{a}, K(\xi)\%),\\
0, & \text{otherwise},
\end{cases}
\end{equation}
where $\mathrm{sort}(\bm{a}, K(\xi)\%)$ returns the indices of the largest $K(\xi)\%$ elements in $\bm{a}$~\cite{ref22}, and the retention ratio $K(\xi)$ is adaptively determined by
\begin{equation}
K(\xi) = 100 \cdot \mathrm{sigmoid}(\gamma \xi + \delta \sigma^2),
\end{equation}
where $\mathrm{sigmoid}(\cdot)$ maps the learnable threshold to $(0,1)$, $\gamma$ is a scaling factor controlling the sensitivity, $\delta$ is a hyperparameter, and $\delta \cdot \sigma^2$ adjusts the threshold according to the channel noise variance, retaining more units under high noise and pruning more under low noise.

Since M generated by selecting the $K(\xi)\%$ units is non-differentiable, we employ the Straight-Through Estimator (STE) during backpropagation. This allows gradients to flow to both the importance scores $\bm{a}$ and the threshold $\xi$.

Let $\mathcal{W}_{\mathrm{pruned}} = \{ \mathbf{W}_{\mathrm{pruned}}(i,j) \}$
denote the set of pruned weight matrices across all  Transformer encoder layers, where \(i\) indexes the layer and \(j\) indexes prunable submodules within that layer. Then the task loss explicitly depends on all pruned weights:
\begin{equation}
\mathcal{L}
= \mathcal{L}_{\mathrm{MSE}}\big(\hat{\mathbf{S}}(\mathcal{W}_{\mathrm{pruned}}),\, \mathbf{S}\big),
\end{equation}
where $\mathbf{S}$ denotes a batch of input training samples, and 
$\hat{\mathbf{S}}(\mathcal{W}_{\mathrm{pruned}})$ denotes the corresponding
reconstructed outputs.

\begin{figure}[t]
    \centering
    \includegraphics[clip, trim=0.1cm 0.1cm 0.1cm 0.1cm, height=2.75cm, width=\linewidth]{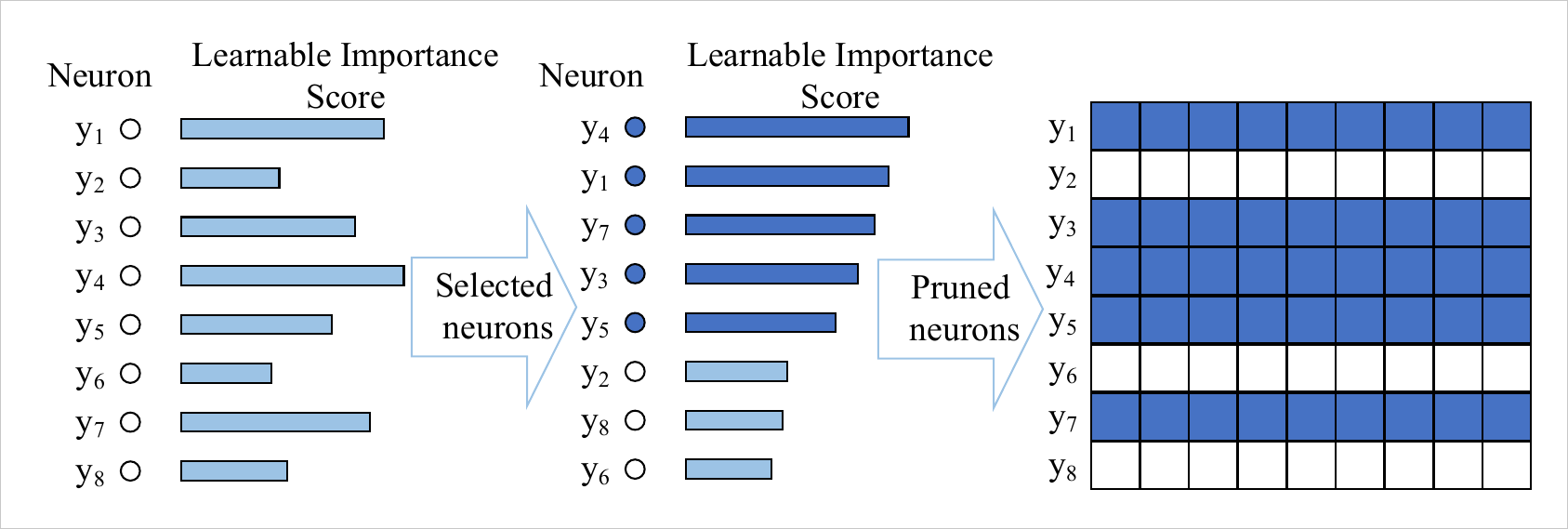} % 修改为你的图片路径
    \caption{Visualization of pruning the fully connected layer in the Transformer encoder layer. Blue neurons are active, white neurons are pruned.}
    \label{fig:FFC_pruning}
\end{figure}

To avoid trivial solutions where pruning is minimized and all units are retained, we introduce a sparsity regularization term $\mathcal{L}_{\rho}$ in the loss function, which penalizes cases where the overall pruning ratio $\bar{\rho}$ falls below a user-defined target ratio $\rho_t$:
\begin{equation}
    \mathcal{L}_{\rho} = \begin{cases}
(\bar{\rho} - \rho_t)^2, & \bar{\rho} < \rho_t, \\
0, & \text{otherwise}.
\end{cases}
\end{equation}

The overall pruning ratio $\bar{\rho}$ is computed as follows. Let $\rho_{i,j} \in [0,1]$ denote the pruning ratio of the $j$-th prunable component in the $i$-th layer, related to the threshold $K(\xi_{i,j})$ by
\begin{equation}
\rho_{i,j} = 1 - \frac{K(\xi_{i,j})}{100}.
\end{equation}

Let $\theta_{i,j}$ denote the number of parameters in component $(i,j)$, and $\theta_i = \sum_j \theta_{i,j}$ the total parameters in layer $i$.  
The effective pruning ratio for layer $i$ is the parameter-weighted average of its components:
\begin{equation}
\rho_i = \frac{\sum_j \theta_{i,j} \, \rho_{i,j}}{\theta_i}.
\end{equation}

The overall pruning ratio across all layers is then
\begin{equation}
\bar{\rho} = \frac{\sum_{i=1}^{L} \theta_i \, \rho_i}{\sum_{i=1}^{L} \theta_i} = \frac{\sum_{i=1}^{L} \sum_j \theta_{i,j} \, \rho_{i,j}}{\Theta_\text{total}},
\end{equation}
where $\Theta_\text{total} = \sum_{i=1}^{L} \theta_i$.

Finally, the total loss combines the reconstruction loss and the sparsity regularization:
\begin{equation}
\mathcal{L} = \mathcal{L}_{\mathrm{MSE}}\big(\hat{\mathbf{S}}(\mathcal{W}_{\mathrm{pruned}}),\, \mathbf{S}\big) + \lambda_{\rho} \, \mathcal{L}_{\rho},
\end{equation}
where $\lambda_{\rho}$ controls the relative weight of the sparsity regularization term $\mathcal{L}_{\rho}$ in the total loss. Following~\cite{ref23}, we adopt the adaptive penalty parameter method, in which $\lambda_{\rho}$ is adjusted as
\begin{equation}
    \lambda_{\rho} = \max \left( \lambda_{max} \frac{\mathcal{L}_{\rho}}{(1-\rho_{\text{t}})^2}, \lambda_{min} \right),
\end{equation}
where $\lambda_{\max}$ and $\lambda_{\min}$ are predefined upper and lower bounds, which ensure that the penalty coefficient $\lambda_\rho$ varies within a reasonable range.

\begin{figure}[t]
    \centering
    \includegraphics[clip, trim=0.1cm 0.1cm 0.1cm 0.1cm, height=2.75cm, width=\linewidth]{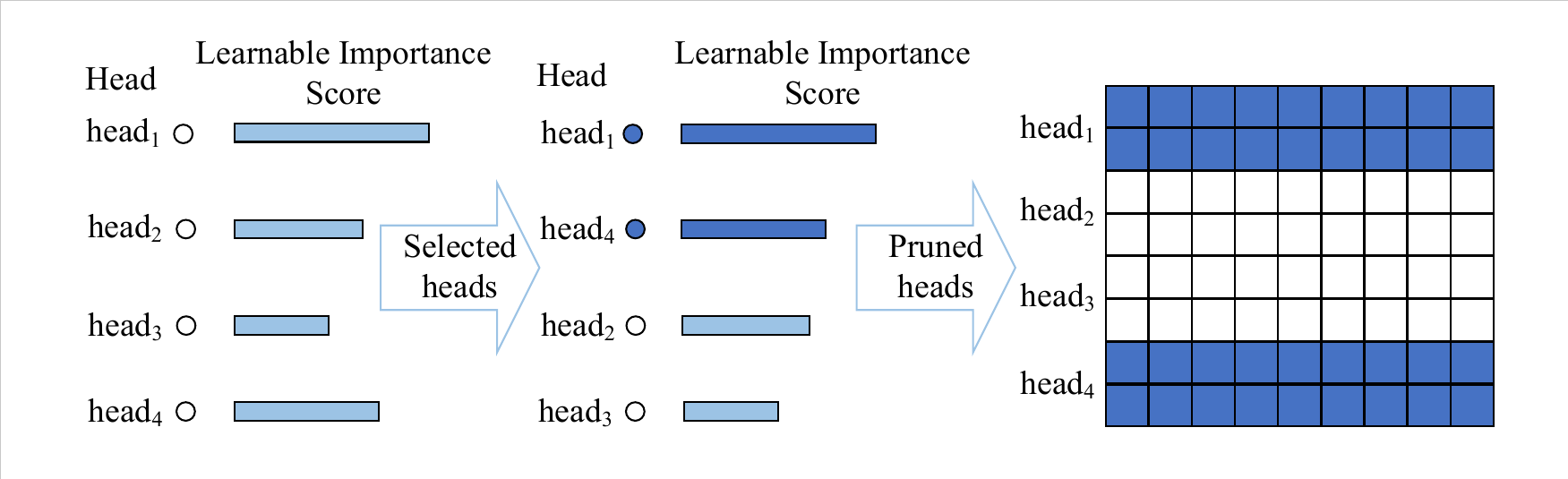} % 修改为你的图片路径
    \caption{Visualization of pruning the MHSA module in the Transformer encoder layer. Blue heads are active, white heads are pruned.}
    \label{fig:mhsa_pruning}
\end{figure}

Fig.~\ref{fig:FFC_pruning} illustrates the pruning process, where neurons are ranked by their importance scores and retained according to the learnable thresholds. Similarly, a subset of attention heads in the MHSA module is selectively activated, as shown in Fig.~\ref{fig:mhsa_pruning}.

\subsection{Joint Width–Depth Optimization Strategy}

The aforementioned strategies treat depth and width as independent dimensions. Based on these two mechanisms, we develop a joint width–depth optimization strategy to further extend the flexibility of the encoder and decoder.

\textbf{Step 1 — Unified width--depth space.}  
Width pruning and depth control together define a two-dimensional structural space
\[
\mathcal{C} = \{ (\rho_t, T) \},
\]
where $\rho_t$ recalls the user-defined target pruning ratio introduced in the width adjustment module, 
and $T$ is the stopping threshold used in the dynamic depth mechanism.
This space characterizes all feasible width--depth configurations that the encoder and decoder can adopt.

\textbf{Step 2 — Width-first Adaptation.}  
Width adjustment is guided by global importance statistics aggregated over many samples, which requires a stable structure for reliable evaluation of neurons and attention heads.  
Depth adjustment, in contrast, is input-dependent, as each image patch may stop at different layers.  
If depth control is applied before width pruning, patch-level variability prevents accurate assessment of unit importance, making width pruning unreliable.  
Therefore, width adjustment is performed first to establish a stable width configuration, followed by dynamic depth adjustment on this stabilized structure.

\textbf{Step 3 — Depth control on the stabilized structure.}  
After stabilizing the width configuration and freezing all width-related parameters, the stopping-score mechanism is trained to achieve input-adaptive depth control.  
During this process, the channel-dependent term $\beta\sigma^{2}$ in the stopping-score computation (Eq.~\eqref{stopsores}) is removed, since the effect of channel conditions has already been accounted for in the width-pruning stage.

\textbf{Step 4 — Width–depth trade-off evaluation.}  
Once both width pruning and depth control are trained, different $(\rho_t, T)$ pairs are evaluated in terms of their reconstruction quality and computational complexity.
Devices with abundant resources may adopt wider and deeper configurations for higher quality, whereas resource-constrained devices select lightweight configurations for real-time operation.

This joint procedure integrates width and depth adaptation with minimal modifications to the original mechanisms, avoiding conflicts that arise when they are optimized separately and resulting in a coherent and flexible structural adjustment framework. The detailed training process is summarized in Algorithm~\ref{alg:joint}.

\begin{algorithm}[t]
\caption{Training procedure for joint width–depth optimization strategy}
\label{alg:joint}
Initialize: target pruning ratio list $\mathcal{R}_w$, stopping threshold list $\mathcal{R}_d$;
\BlankLine
\For{$\rho_t \in \mathcal{R}_w$}{
    Perform width pruning for target pruning ratio $\rho_t$\;
    Obtain the pruned backbone\;
    \For{$T \in \mathcal{R}_d$}{
        Freeze all width-related parameters\;
        Fine-tune the depth controller on the frozen width-pruned backbone using stopping threshold $T$\;
        Save the final configuration\;
    }
}
\end{algorithm}

\section{Experiment}

\subsection{Simulation Settings and Baselines}
We conduct simulations on the CIFAR-10 dataset to evaluate the proposed semantic communication system and its dynamic adjustment strategies with image reconstruction as the target task. We employ the Structural Similarity Index Measure (SSIM)~\cite{ref24} as the evaluation metric to assess the visual fidelity between the reconstructed images and their original counterparts. For clarity, the recursive ViT–based backbone is denoted as RecViT, while the dynamic depth, dynamic width, and joint width–depth optimization strategies are referred to as Proposed (DD), Proposed (DW), and Proposed (Joint), respectively. The parameter settings for the encoder and decoder are summarized in Table~\ref{tab:sim_params}.

For comparison purposes, we consider several
classic and state-of-the-art baselines, which are:

\begin{table}[t]
\centering
\caption{Simulation Parameter Settings}
\label{tab:sim_params}
\begin{tabular}{l l l}
\hline
\textbf{Network Layer Name} & \textbf{Parameter Type} & \textbf{Value} \\
\hline
Recursive Transformer Unit & Number & 4 \\
& Image Patch Size & 8 \\
& Embedding Dimension & 192 \\
& Number of Attention Heads & 8 \\
& MLP Multiplier & 4 \\
& Activation Function & ReLU \\
Linear Layer & Number of Neurons & 64 \\
Dynamic Depth Module & Number of Neurons & 192 \\
& Activation Function & sigmoid \\

\hline
\end{tabular}
\end{table}
\begin{figure}[t]
    \centering
    \subfloat[AWGN channel]{%
        \includegraphics[width=0.48\textwidth]{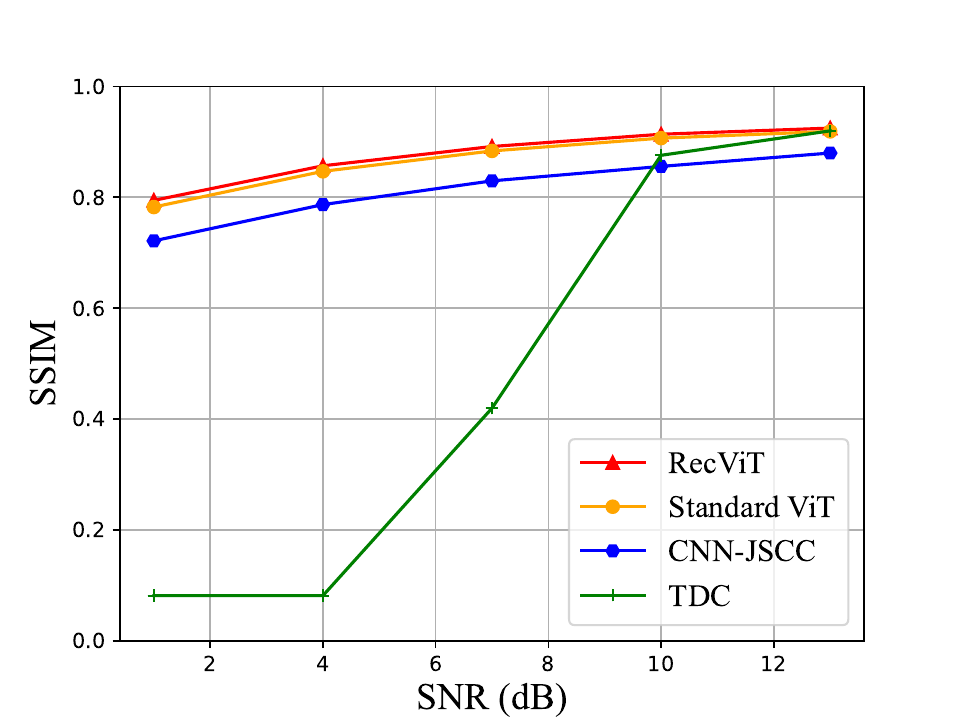}
        \label{fig:4-1a}
    }
    \hfill
    \subfloat[Rayleigh channel]{%
        \includegraphics[width=0.48\textwidth]{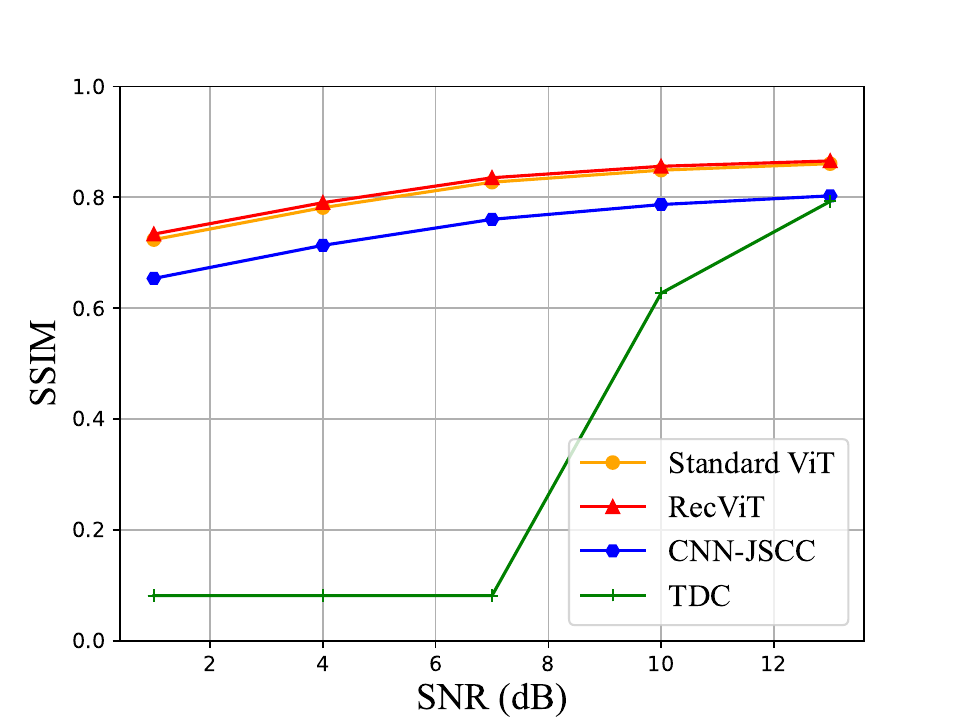}
        \label{fig:4-1b}
    }

    \caption{SSIM performance varies with SNR under different channel types.}
    \label{fig:4-1}
\end{figure}

\begin{enumerate}
\item \textbf{Standard ViT}~\cite{ref25}:
Standard ViT uses eight Transformer encoder layers to extract semantic features and reconstruct images.

\item \textbf{CNN-JSCC}~\cite{ref26}:  
CNN-JSCC exploits local spatial correlations in images to extract semantic features for end-to-end transmission and reconstruction.

\item \textbf{TDC}:  
TDC combines BPG compression, LDPC channel coding, and 16-QAM modulation for reliable image transmission.

\item \textbf{RecViT-EE}~\cite{ref27}: 
RecViT-EE applies an early-exit mechanism to RecViT, determining the number of executed layers using a pre-defined exit criterion.

\item \textbf{RecViT-SP:}~\cite{ref28}  
RecViT-SP applies a static pruning strategy to RecViT, removing a fixed proportion of neurons and attention heads in advance.

\item \textbf{RecViT-SDW}~\cite{ref29}: RecViT-SDW applies a fixed dual-dimension strategy to RecViT, adjusting both depth and width using constant layer numbers and predetermined pruning ratios.
\end{enumerate}

\subsection{SSIM Performance Analysis}

\begin{figure*}[t]
    \centering

    \subfloat[Average SSIM difference–FLOPs across different $(\rho_t, T)$ settings.]{%
        \includegraphics[width=0.48\textwidth]{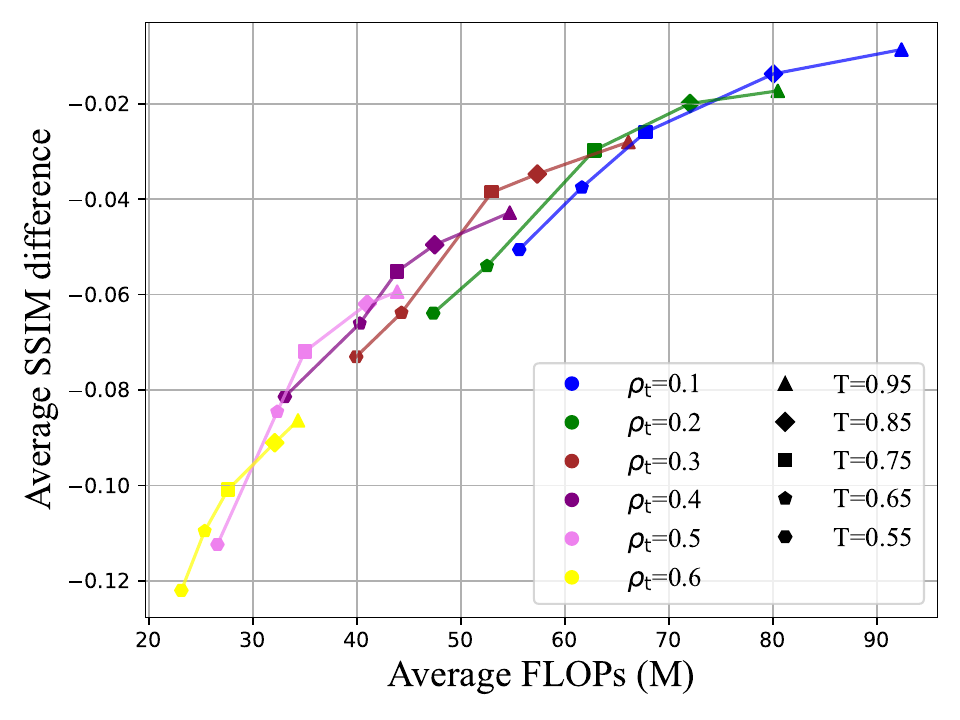}
        \label{fig:4-12a}
    }
    \hfill
    \subfloat[Optimal width–depth configuration achieving the best average SSIM difference–FLOPs balance.]{%
        \includegraphics[width=0.48\textwidth]{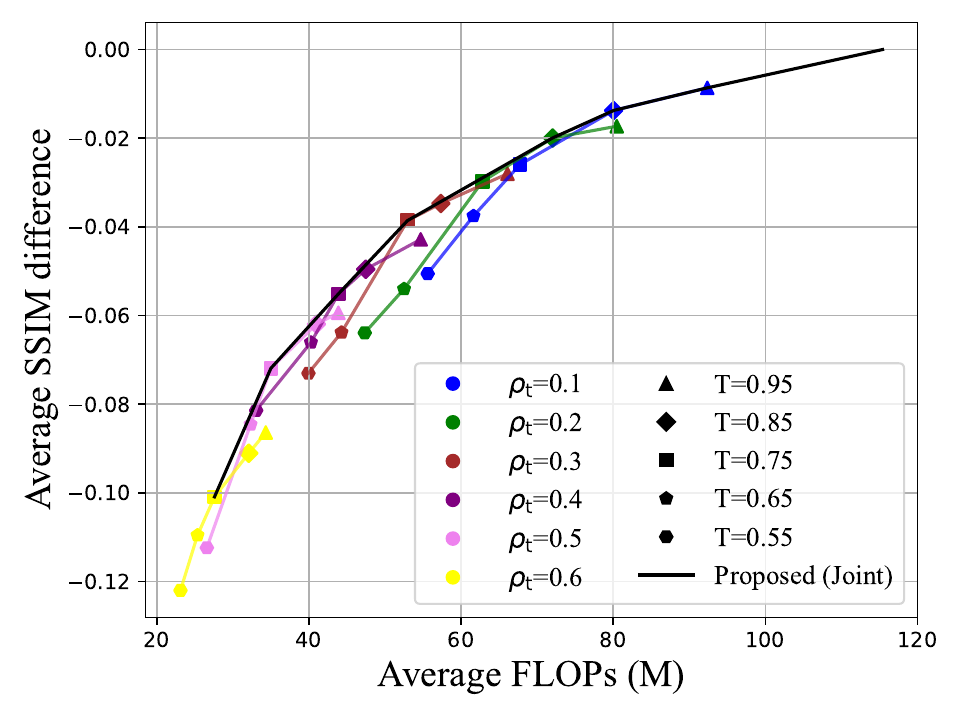}
        \label{fig:4-12b}
    }

    \caption{Average SSIM difference (relative to Standard ViT) vs. average FLOPs under the Rayleigh channel for the joint width–depth optimization strategy.}
    \label{fig:4-12}
\end{figure*}

\begin{figure*}[t]
    \centering

    \subfloat[AWGN channel]{%
        \includegraphics[width=0.48\textwidth]{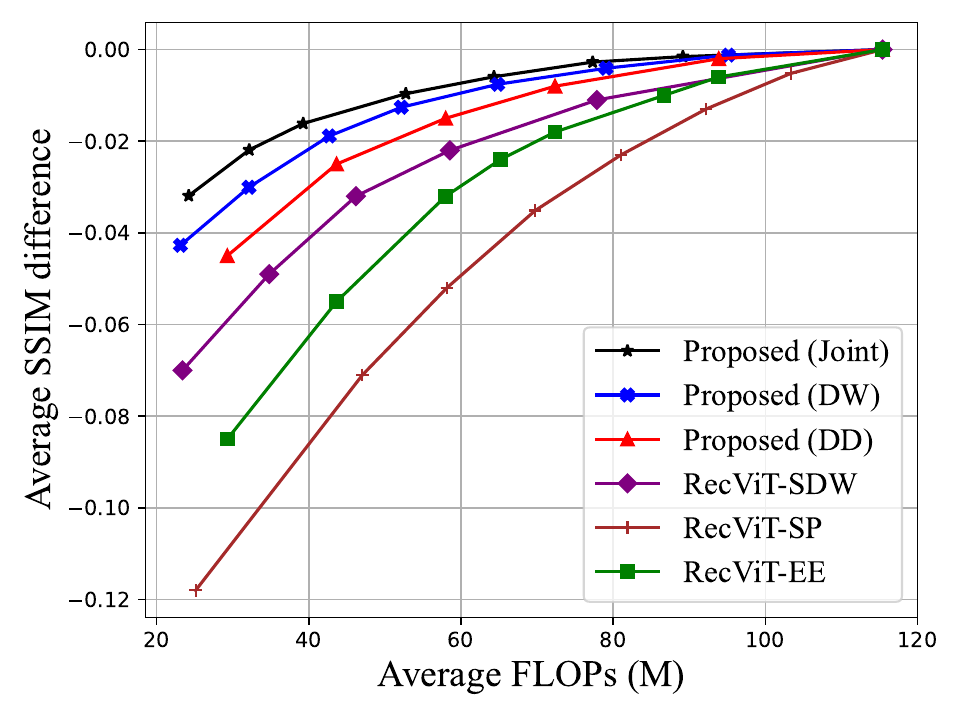}
        \label{fig:4-15a}
    }
    \hfill
    \subfloat[Rayleigh channel]{%
        \includegraphics[width=0.48\textwidth]{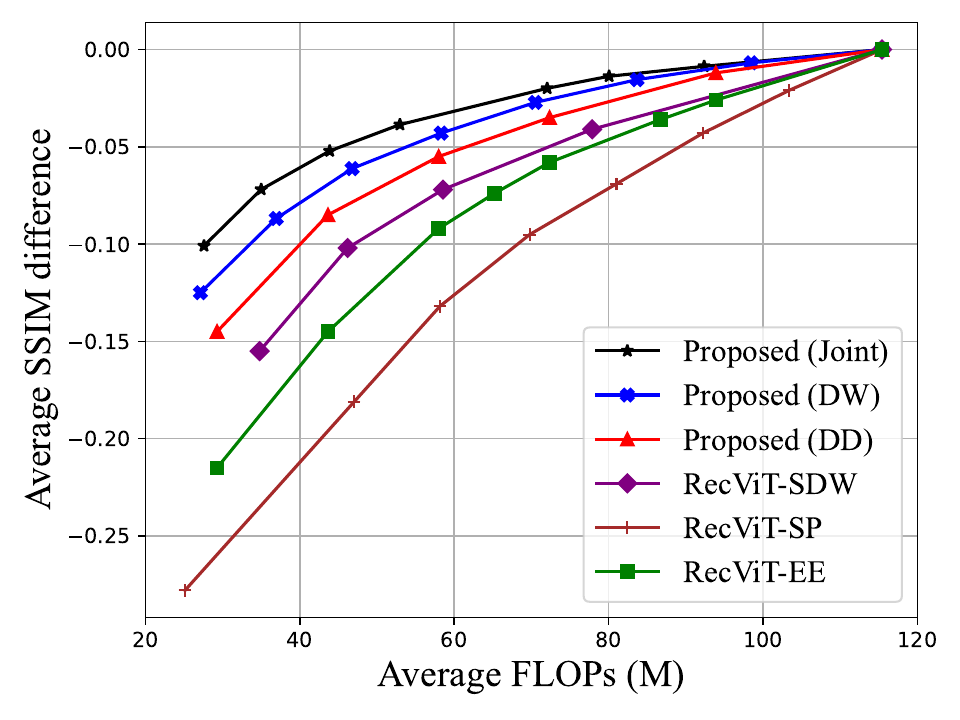}
        \label{fig:4-15b}
    }

    \caption{Average SSIM difference vs. average FLOPs for all considered algorithms under different channel types.}
    \label{fig:4-15}
\end{figure*}

Fig.~\ref{fig:4-1} shows how the SSIM performance of all considered algorithms varies as SNR increases under the AWGN and Rayleigh channels. The proposed RecViT consistently achieves higher SSIM than Standard ViT, as it recursively reuses Transformer encoder layers, reducing the total number of parameters and improving learning efficiency. Both ViT-based systems outperform CNN-JSCC, as the self-attention mechanism in Transformers enables a wider receptive field and more effective modeling of long-range dependencies in images, allowing semantic features to be better captured and reducing quality degradation under noisy channels. In contrast, the TDC scheme is highly sensitive to channel noise and suffers severe SSIM collapse at low SNRs, highlighting the robustness advantage of semantic communication.

Fig.~\ref{fig:4-12} visualizes the landscape of average SSIM difference relative to Standard ViT and average FLOPs for the proposed joint width–depth optimization strategy under the Rayleigh channel. In Fig.~\ref{fig:4-12a}, each point corresponds to a specific $(\rho_t, T)$ pair. The results reveal a clear trade-off between the two dimensions: adjusting pruning ratios and stopping thresholds jointly influences image reconstruction quality. For instance, a configuration with 20\% pruning and an average depth of 12 layers at approximately 60M FLOPs achieves higher SSIM performance than a configuration with 10\% pruning and an average depth of 11 layers. Based on these observations, we select the width–depth configurations that achieve the best average SSIM difference–FLOPs balance, as shown in Fig.~\ref{fig:4-12b}.

\begin{figure*}[t]
    \centering

    \subfloat[SSIM vs. SNR at 30M FLOPs under the AWGN channel.]{%
        \includegraphics[width=0.48\textwidth]{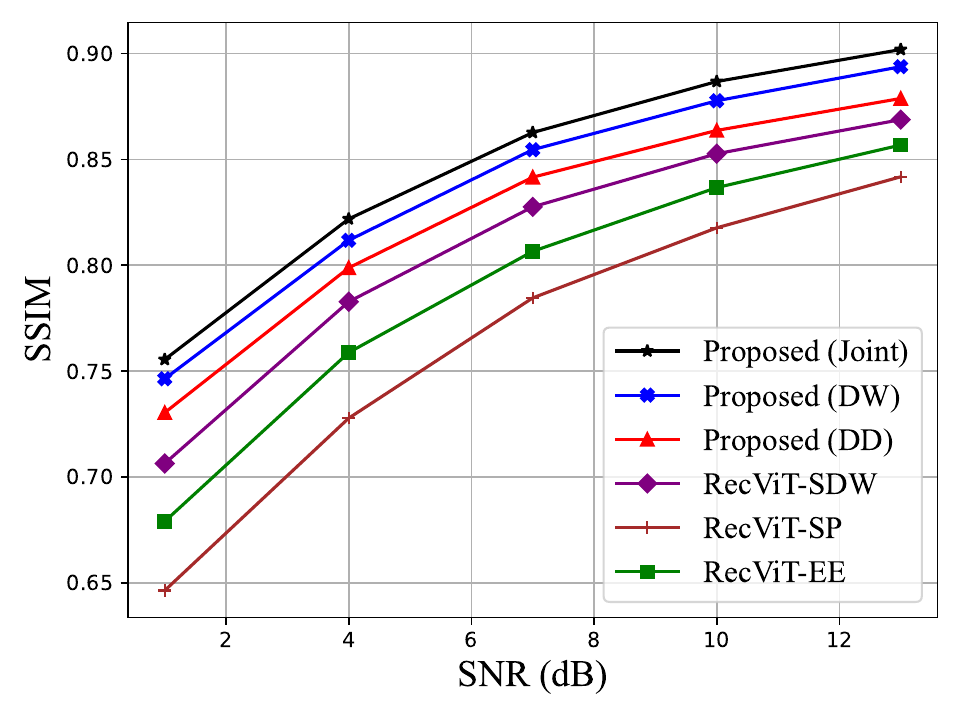}
        \label{fig:4-17a}
    }
    \hfill
    \subfloat[SSIM vs. SNR at 40M FLOPs under the Rayleigh channel.]{%
        \includegraphics[width=0.48\textwidth]{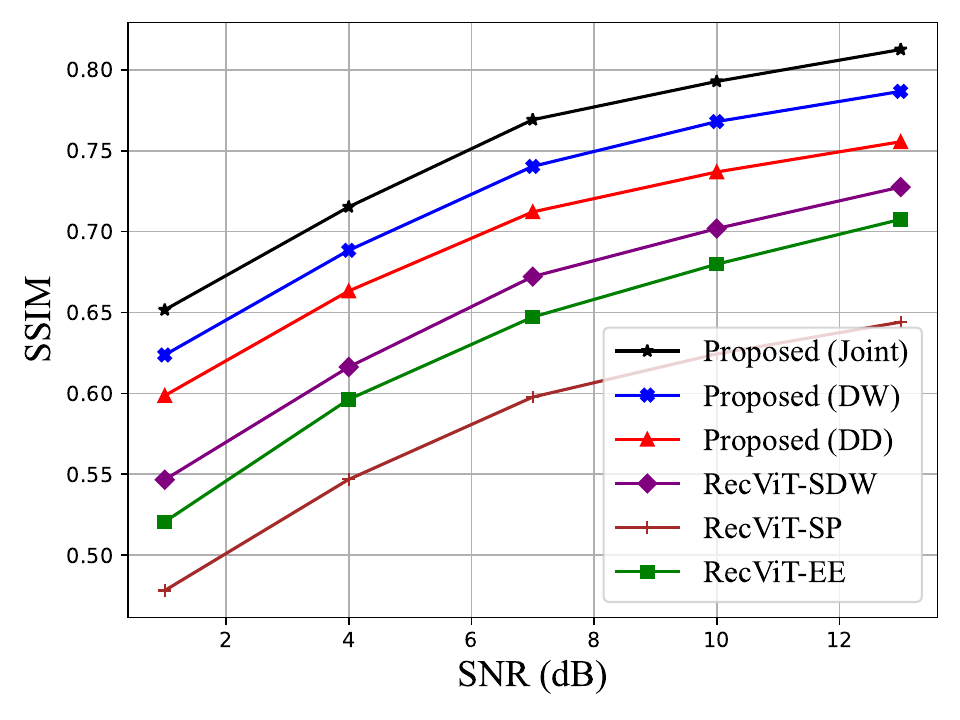}
        \label{fig:4-17b}
    }

    \caption{SSIM performance as SNR increases at fixed computational complexity under different channel types.}
    \label{fig:4-17}
\end{figure*}

\begin{figure}[t]
    \centering
    \includegraphics[width=0.48\textwidth]{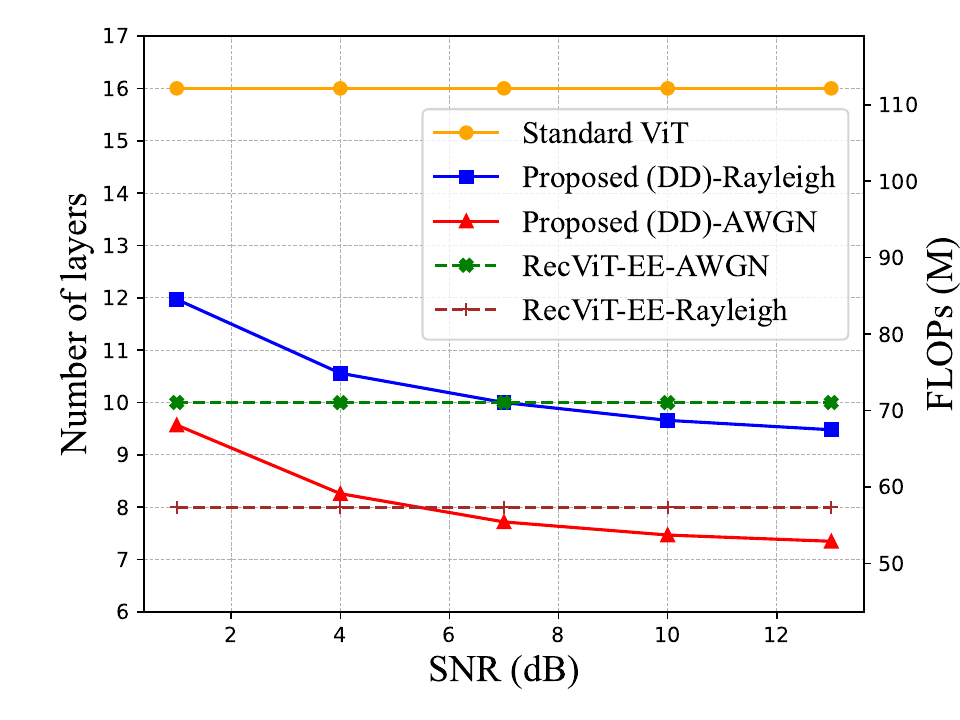}
    \caption{The number of layers and the corresponding FLOPs vary as SNR increases for the proposed dynamic depth adjustment strategy, with Standard ViT and RecViT-EE as references.}
    \label{fig:depth_comp}
\end{figure}

\begin{figure}[t]
    \centering
    \includegraphics[width=0.48\textwidth]{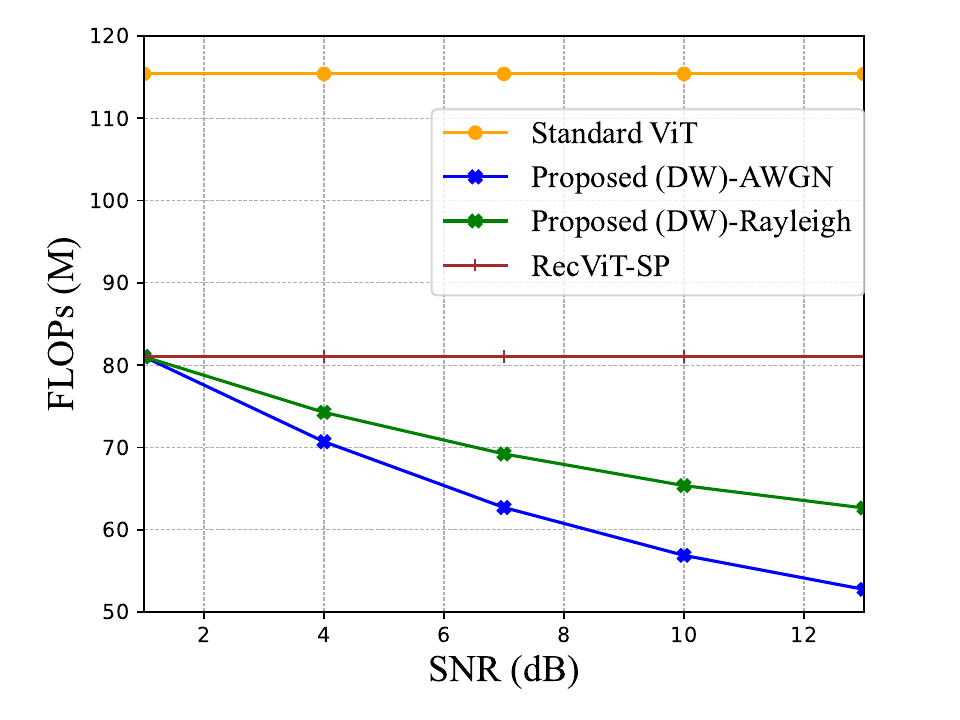}
    \caption{
    The FLOPs vary as SNR increases for  the proposed dynamic width adjustment strategy, with Standard ViT and RecViT-SP as references..}
    \label{fig:comp_width_awgn}
\end{figure}

Fig.~\ref{fig:4-15} shows the landscape of the average SSIM difference and the average FLOPs for all considered algorithms. 
At the same computational complexity, the proposed joint width–depth optimization strategy exhibits the smallest performance degradation. 
For example, when the computational complexity is reduced to about 30\% of Standard ViT, 
the proposed joint width–depth optimization strategy incurs an SSIM drop of only 2\% under the AWGN channel, 
which is smaller than the drops of 3\% and 4\% for the proposed dynamic width adjustment and dynamic depth adjustment strategies, as shown in Fig.~\ref{fig:4-15a}. 
Under the Rayleigh channel, the proposed joint width–depth optimization strategy yields an average SSIM difference of about 7\%, 
compared with 9\% for the proposed dynamic width adjustment strategy and 11\% for the proposed dynamic depth adjustment strategy, as shown in Fig.~\ref{fig:4-15b}. 
Overall, the proposed dynamic strategies outperform the baseline schemes at equal computational complexity, and the proposed joint width–depth optimization strategy 
provides the most favorable trade-off between image reconstruction quality and computational efficiency.

Fig.~\ref{fig:4-17} shows how SSIM varies as SNR increases
under a fixed computational complexity for all considered algorithms, with the proposed joint width–depth optimization strategy 
consistently maintaining high SSIM across all SNR levels. Fig.~\ref{fig:4-17a} 
shows that under the AWGN channel, where the computational complexity is approximately 
30M FLOPs, the proposed joint width–depth optimization strategy incurs SSIM reductions of about 2.7\% at 1 dB 
and 1.7\% at 13 dB relative to Standard ViT. 
Under the Rayleigh channel, Fig.~\ref{fig:4-17b} shows that with a computational complexity of around 40M FLOPs, the proposed joint width–depth optimization strategy incurs SSIM reductions of 7.2\% at 1 dB and 4.8\% at 13 dB.

\subsection{Parameter Count and Adaptive Computational complexity}

\subsubsection{Recursive ViT Parameter Count}
Table~\ref{tab:param_count} compares the number of parameters of different backbone designs. Owing to its weight-sharing recursive structure, RecViT requires only 3.70M parameters, achieving a reduction of nearly 48.7\% compared with Standard ViT.

\begin{table}[t]
\centering
\caption{Parameter Count Comparison}
\label{tab:param_count}
\begin{tabular}{l c}
\hline
\textbf{System} & \textbf{Number of Parameters (Million)} \\
\hline
ViT & 7.23 \\
RecViT & 3.70 \\
\hline
\end{tabular}
\end{table}
\begin{figure}[t]
    \centering
    \includegraphics[width=0.48\textwidth]{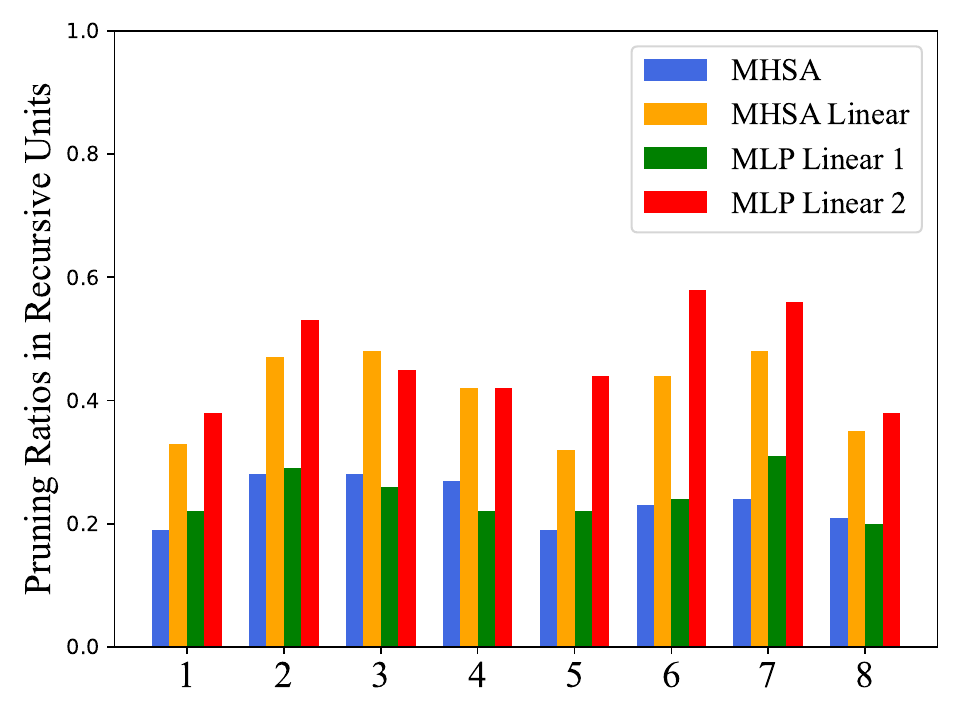}
    \caption{Distribution of pruning ratios for the RTUs under the AWGN channel when SNR $=$ 1 dB.}
    \label{fig:pruning_ratio}
\end{figure}
\subsubsection{Adaptive Computational Complexity Analysis}

Fig.~\ref{fig:depth_comp} shows how the proposed dynamic depth adjustment strategy adapts the number of Transformer encoder layers as SNR increases when $T = 0.75$. As SNR increases, the proposed dynamic depth adjustment strategy progressively skips more encoding–decoding layers, thereby reducing computational Complexity according to the channel condition.
Under the AWGN channel, the average depth is 8.02 layers, yielding a 49.6\% reduction in FLOPs compared with the fixed 16-layer Standard ViT. Under the Rayleigh channel, the average depth increases to 10.29 layers to accommodate channel fluctuation, while still achieving a 35.5\% FLOPs reduction relative to Standard ViT.
In contrast, RecViT-EE operates with a fixed depth across all SNR levels and therefore lacks such adaptive flexibility.

Fig.~\ref{fig:comp_width_awgn} shows how the proposed dynamic width adjustment strategy reduces computational
complexity by dynamically adjusting width as SNR increases when $\rho_t = 30\%$.  
At low SNR of 1 dB, the computational complexity of the proposed dynamic width adjustment strategy is close to that of RecViT-SP, which uses a static pruning ratio of 30\%. This corresponds to a 29.8\% reduction in FLOPs compared with Standard ViT. As SNR increases, the proposed dynamic width adjustment strategy adaptively increases the pruning ratio, lowering the computational complexity. These results indicate that the proposed dynamic width adjustment strategy flexibly adjusts computation according to channel conditions, further enhancing computational efficiency.

Fig.~\ref{fig:pruning_ratio} gives the learned pruning ratios of the eight RTUs under the AWGN channel when SNR $=$ 1 dB. The results demonstrate that the proposed dynamic width adjustment strategy adaptively prunes both neurons and attention heads.

\section{Conclusion}

In this paper, we proposed a recursive ViT–based image semantic communication system with three dynamic adjustment strategies: dynamic depth adjustment, dynamic width pruning, and joint width–depth optimization. The recursive structure and adaptive strategies jointly enabled significant reductions in parameter count and computational complexity. Simulation results verified that the proposed system, combined with the three dynamic adjustment strategies, reduced the parameter count by 48.7\% and achieved higher reconstruction quality than existing baselines under comparable computational complexity.

\bibliographystyle{IEEEtran}  
\bibliography{refs}

\begin{IEEEbiography}[{\includegraphics[width=1in,height=1.25in,clip,keepaspectratio]{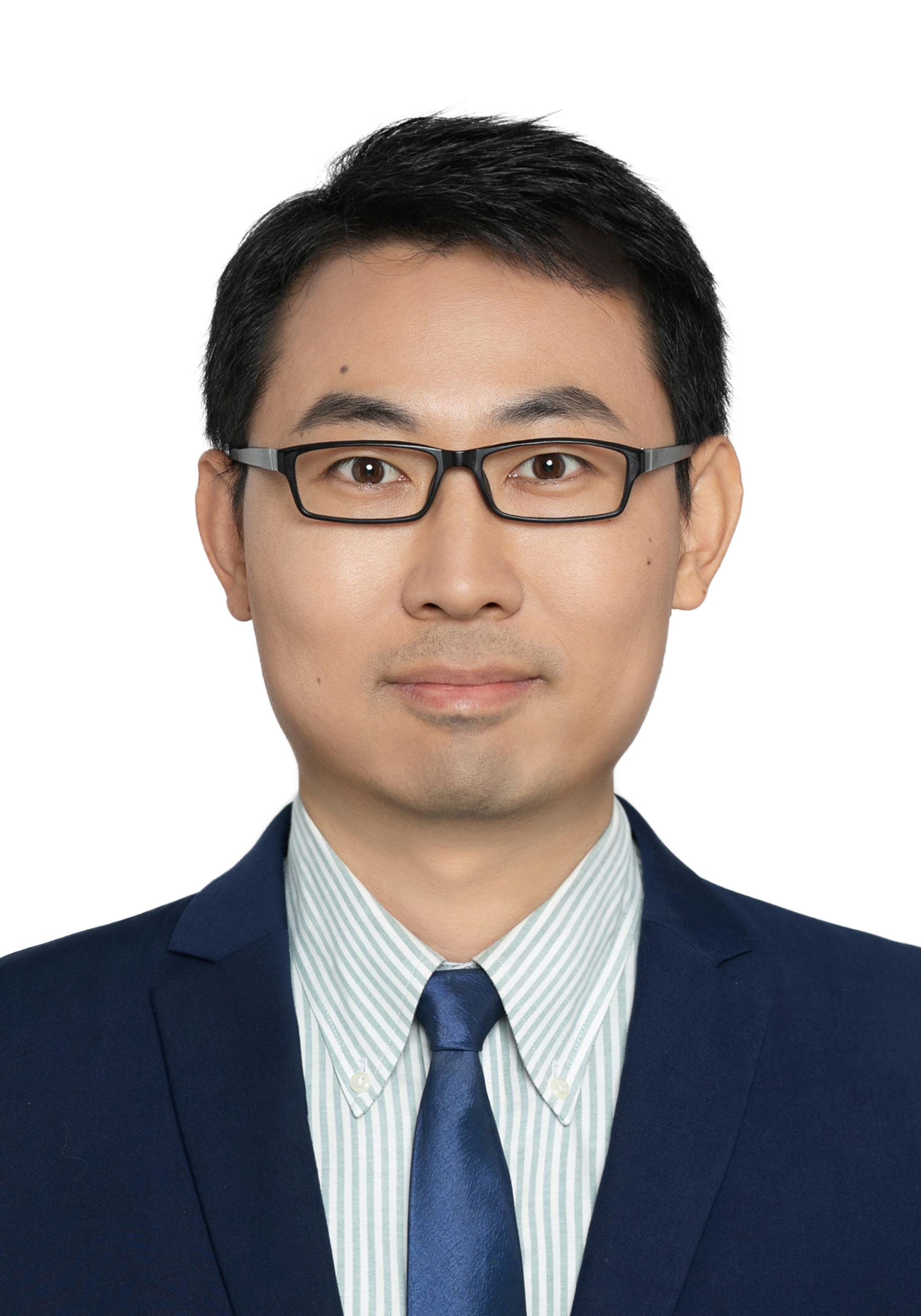}}]{Zhilong Zhang}
 received the B.E. degree in communication engineering from the University of Science and Technology, Beijing, China in 2007, and the M.S. and the Ph.D. degrees in communication and information systems from Beijing University of Posts and Telecommunications (BUPT), Beijing, China in 2010 and 2016, respectively. He is currently an associate professor at BUPT. From 2010 to 2012, he was a software engineer at TD Tech Ltd., Beijing, China. From 2014 to 2015, he was a visiting scholar at Stony Brook University, NY, USA. His research interests include optimization theory and machine learning, and their applications in wireless networks and video transmissions.
\end{IEEEbiography}

\begin{IEEEbiography}[{\includegraphics[width=1in,height=1.25in,clip,keepaspectratio]{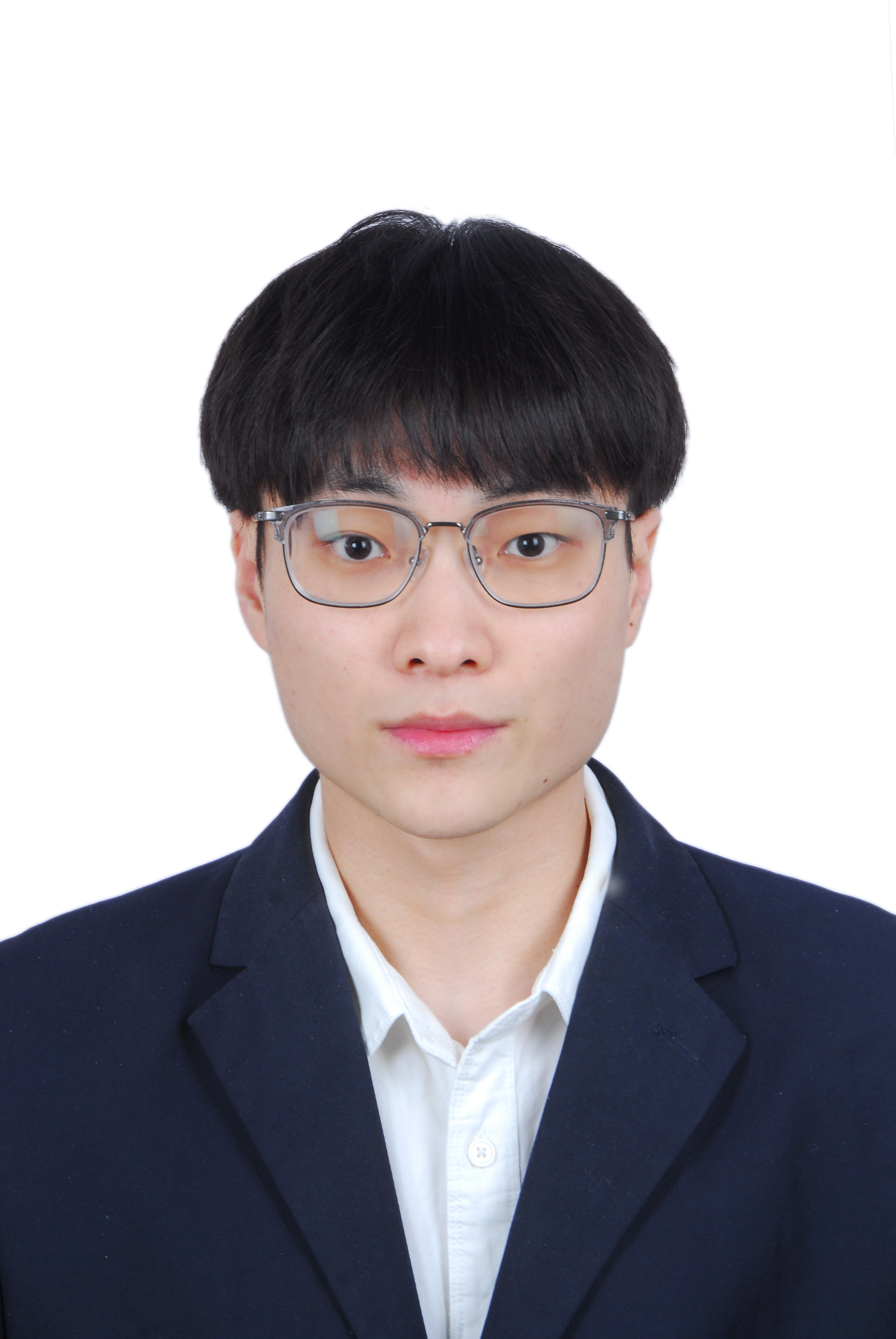}}]{Xinhui Zhang}
received the B.E. degree in information engineering from Beijing University of Posts
and Telecommunications (BUPT), Beijing, China, in
2025. He is currently pursuing the M.S. degree with the Laboratory of Wireless Communication Systems and Networks, BUPT. His main research interests focus on semantic communications.
\end{IEEEbiography}

\begin{IEEEbiography}[{\includegraphics[width=1in,height=1.25in,clip,keepaspectratio]{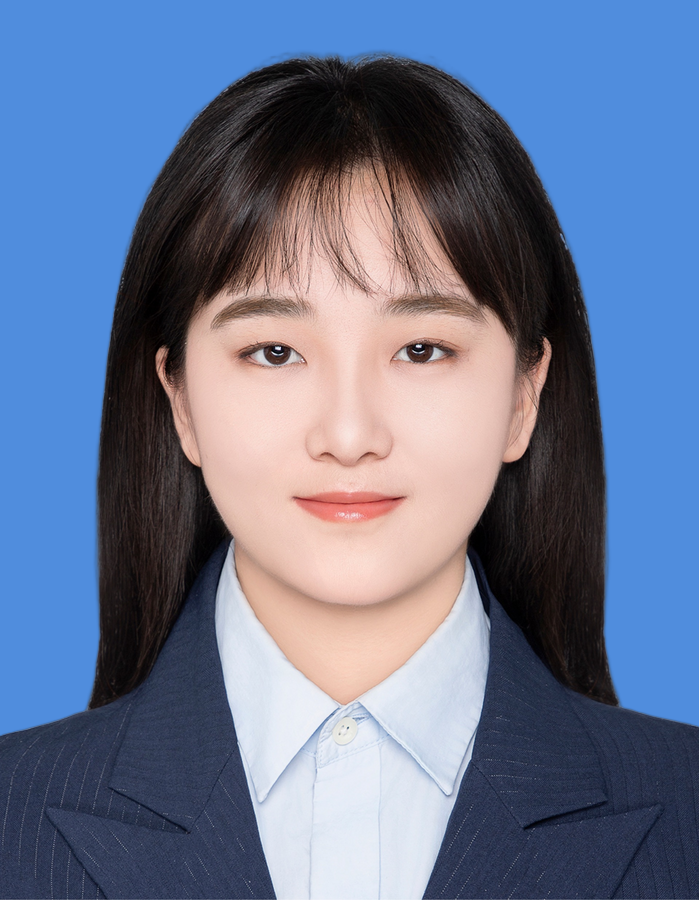}}]{Gongyu Jin}
received the B.E. and M.S. degrees in Communication Engineering from Beijing University of Posts and Telecommunications (BUPT), Beijing, China, in 2021 and 2024. Her research interests include semantic communications and dynamic neural networks.
\end{IEEEbiography}

\begin{IEEEbiography}[{\includegraphics[width=1in,height=1.25in,clip,keepaspectratio]{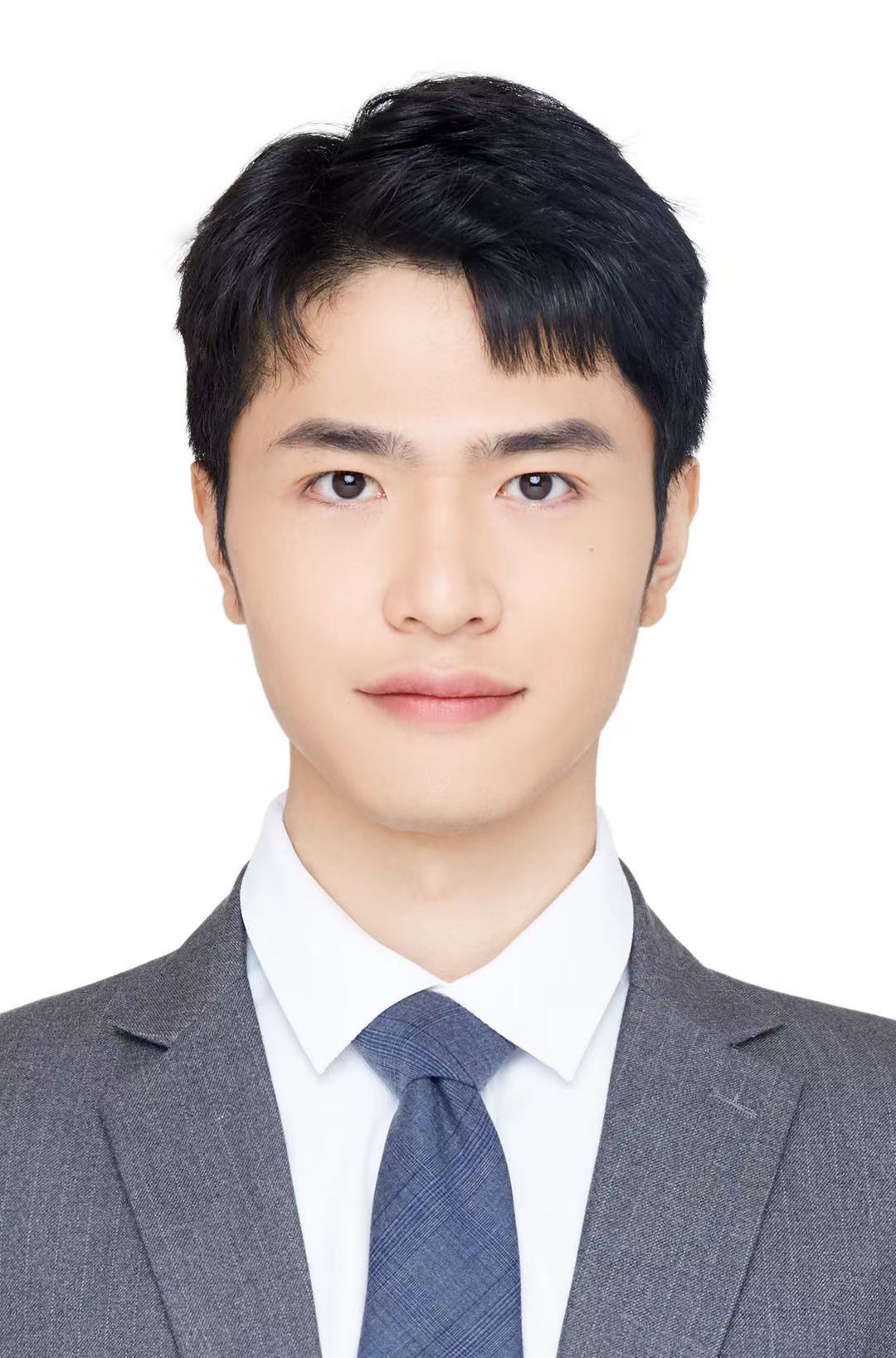}}]{Sihua Wang}
received the Ph.D. degree from the
Beijing University of Posts and Telecommunications
(BUPT) in 2021. He is currently an Assistant Professor with the School of Information and Communication Engineering, BUPT, Beijing, China. From 2021 to 2024, he was a Post-Doctoral Researcher at School of Computer Science (National Pilot Software Engineering School), BUPT. From 2022 to 2024, he held a Hong Kong Scholar Fellow position with the Department of Electronic and Computer Engineering, The Hong Kong University of Science and Technology, Hong Kong. His research interests include mobile edge computing, resource allocation, and machine learning in wireless networks.
\end{IEEEbiography}

\begin{IEEEbiography}[{\includegraphics[width=1in,height=1.25in,clip,keepaspectratio]{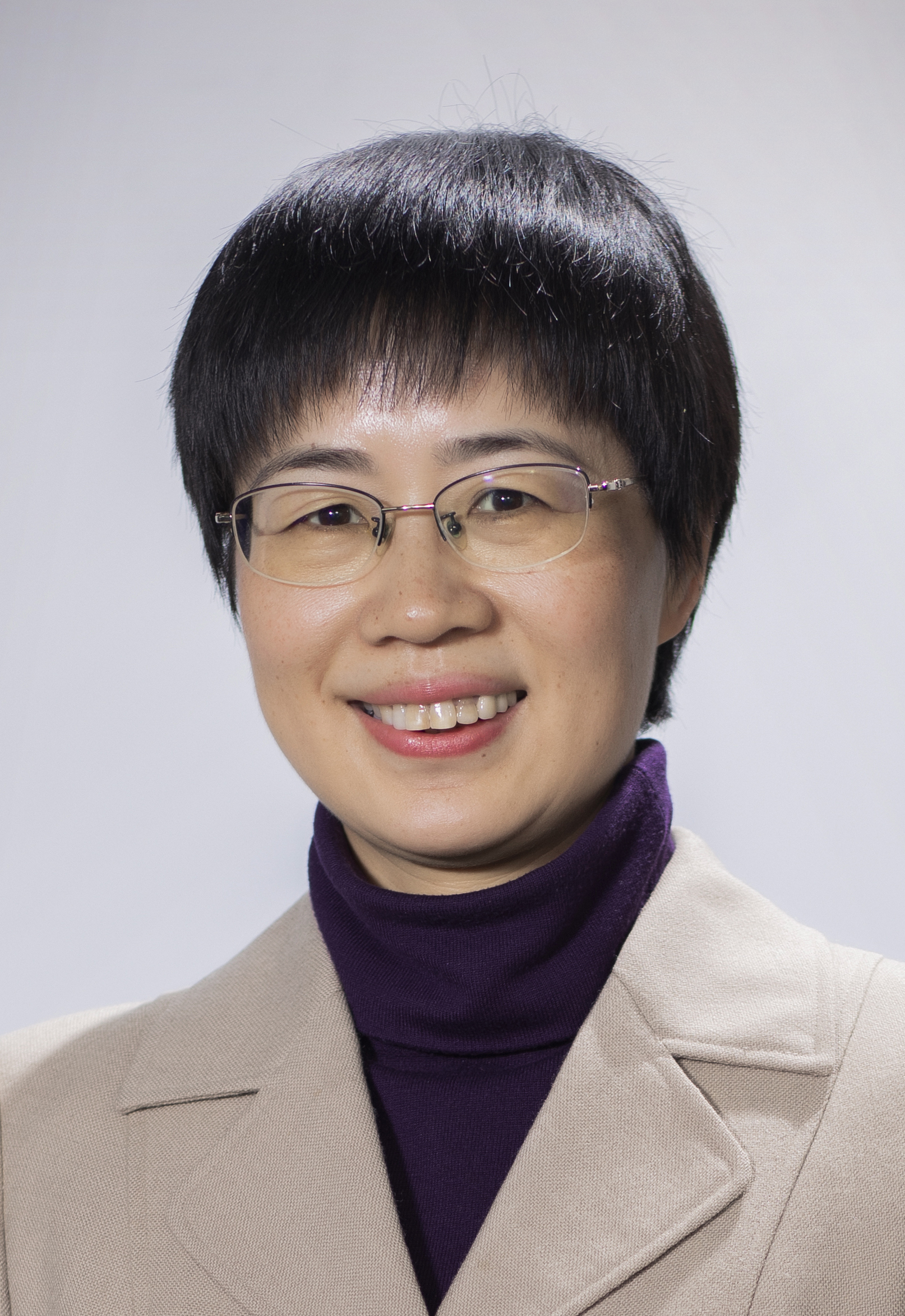}}]{Danpu Liu}
received the Ph.D. degree in communication and electrical systems from Beijing University of Posts and Telecommunications, Beijing, China, in 1998. She was a visiting scholar with the City University of Hong Kong, Hong Kong, in 2002; The University of Manchester, Manchester, U.K., in 2005; and the Georgia Institute of Technology, Atlanta, GA, USA, in 2014. She is currently with the Beijing Key Laboratory of Network System Architecture and Convergence, Beijing University of Posts and Telecommunications. Her current research interests include B5G/6G mobile communications and air–space–ground integrated networks.
\end{IEEEbiography}

\begin{IEEEbiography}[{\includegraphics[width=1in,height=1.25in,clip,keepaspectratio]{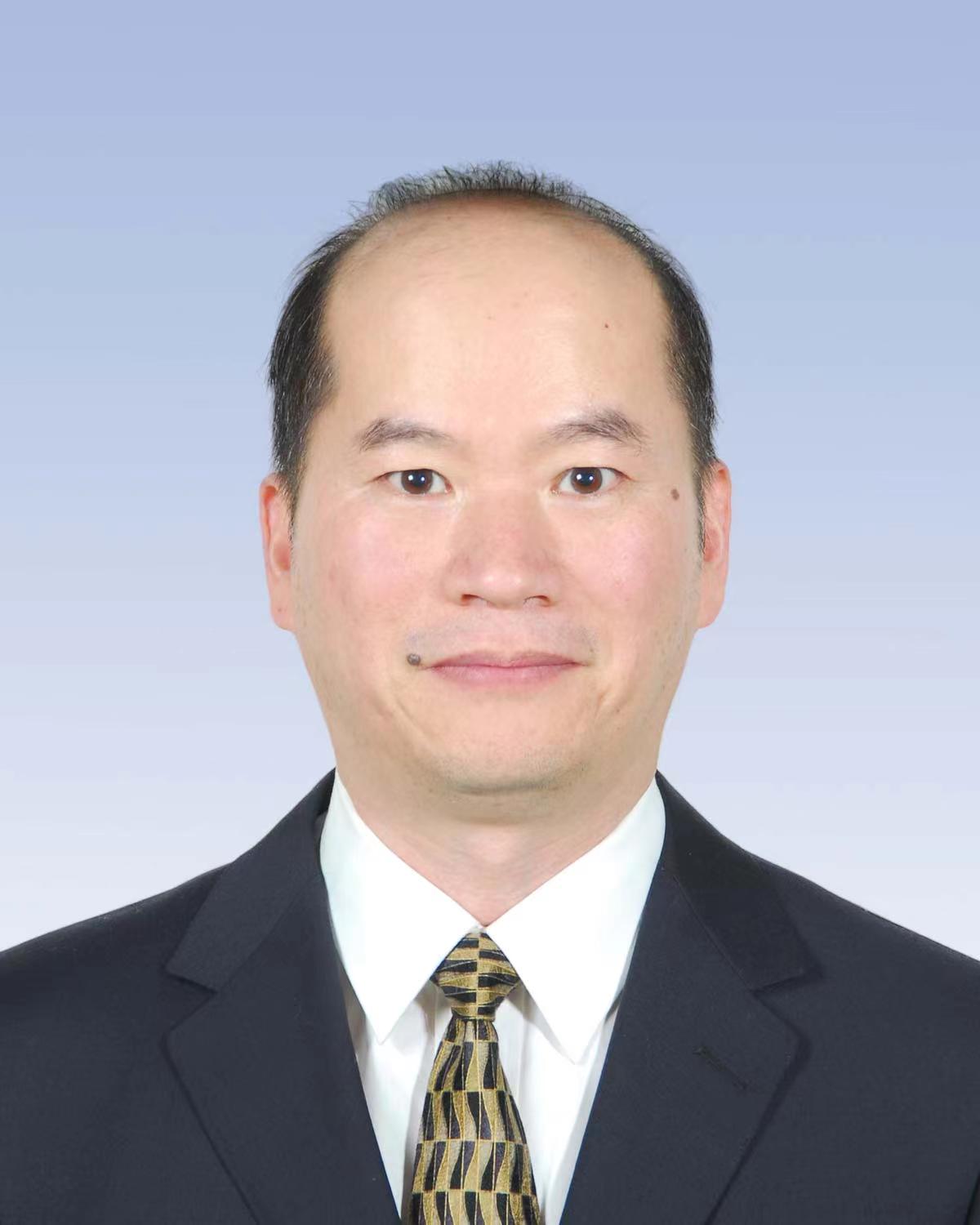}}]{Changchuan Yin}
(M’98–SM’15) received the Ph. D. degree in signal and information processing from Beijing University of Posts and Telecommunications, Beijing, P. R. China, in 1998. In 2004, he was a visiting scholar in the Faculty of Science, the University of Sydney, Sydney, NSW, Australia. From 2007 to 2008, he held a visiting position with the Department of Electrical and Computer Engineering, Texas A\&M University, College Station, TX, USA. He is currently a professor with the School of Information and Communication Engineering, Beijing University of Posts and Telecommunications. His research interests include wireless networks and statistical signal processing. He was the co-recipient of the IEEE Guglielmo Marconi Prize Paper Award in 2023 and the IEEE International Conference on Wireless Communications and Signal Processing Best Paper Award in 2009. He has served as the symposium co-chair and TPC member for many IEEE conferences.
\end{IEEEbiography}
\end{document}